\documentclass[letterpaper, 10 pt, conference]{ieeeconf}  
\IEEEoverridecommandlockouts
\usepackage{cite}
\usepackage{amsmath,amssymb,amsfonts}
\usepackage{algorithmic}
\usepackage{graphicx}
\usepackage[hyphens,spaces,obeyspaces]{url}
\usepackage{hyperref}
\usepackage{todonotes}
\usepackage{textcomp}
\usepackage{xcolor}
\usepackage{bm}
\usepackage{threeparttable}
\usepackage{siunitx}
\usepackage{multirow}
\usepackage{comment}
\usepackage{lipsum}
\usepackage{siunitx}

\def\BibTeX{{\rm B\kern-.05em{\sc i\kern-.025em b}\kern-.08em
    T\kern-.1667em\lower.7ex\hbox{E}\kern-.125emX}}

\hypersetup{
	colorlinks=true,
	linkcolor=blue,
	filecolor=magenta,      
	urlcolor=blue,
}

\renewcommand{\baselinestretch}{0.975}

\begin{document}

\title{\LARGE \bf 
The Magni Human Motion Dataset: Accurate, Complex,\\ Multi-Modal, Natural, Semantically-Rich and Contextualized
}


\author{Tim~Schreiter$^1$,
       Tiago Rodrigues de Almeida$^1$,
       Yufei Zhu$^1$,
       Eduardo Gutierrez Maestro$^1$,\\
       Lucas Morillo-Mendez$^1$,
       Andrey Rudenko$^2$, 
       Tomasz P. Kucner$^3$,
       Oscar Martinez Mozos$^1$,\\
       Martin Magnusson$^1$,
       Luigi Palmieri$^2$,
       Kai O. Arras$^2$,
       and~Achim J.~Lilienthal$^1$
\thanks{$^{1}$
	\"Orebro University, Sweden {\tt\small \{tim.schreiter, tiago.almeida, eduardo.gutierrez-maestro, yufei.zhu, lucas.morillo, oscar.mozos, martin.magnusson, achim.lilienthal\}@oru.se}}
\thanks{$^{2}$Robert Bosch GmbH, Corporate Research, Stuttgart, Germany
{\tt\small \{andrey.rudenko, luigi.palmieri, kaioliver.arras\}@de.bosch.com}}%
\thanks{$^{3}$Mobile Robotics Group, Department of Electrical Engineering and Automation, Aalto University, Finland
{\tt\small tomasz.kucner@aalto.fi}}%
\thanks{This work was supported by the European Union’s Horizon 2020 research and innovation program under grant agreement No. 101017274 (DARKO) and the Wallenberg AI, Autonomous
Systems and Software Program (WASP) funded by the Knut and Alice Wallenberg Foundation.}}

\maketitle

\begin{abstract}
    Rapid development of social robots stimulates active research in human motion modeling, interpretation and prediction, proactive collision avoidance, human-robot interaction and co-habitation in shared spaces. Modern approaches to this end require high quality datasets of human motion trajectories for training and evaluation.
    However, the majority of  available datasets suffers  from either inaccurate tracking data or unnatural, scripted behavior of the tracked people.
    This paper attempts to fill this gap by providing high quality tracking information from motion capture, eye-gaze trackers and on-board robot sensors in a semantically-rich environment.
    To induce natural behavior of the recorded participants, we utilise loosely scripted task assignment, which induces the participants to navigate through the dynamic laboratory environment in a natural and purposeful way towards the randomly assigned targets. The motion dataset, presented in this paper, sets a high quality standard, as the realistic and accurate data is enhanced with semantic information, enabling development of new algorithms which rely not only on the tracking information but also on contextual cues of the moving agents, static and dynamic environment.
\end{abstract}

\section{Introduction}\label{sec:related}

In recent years, the topics of human motion prediction and human-robot interaction have been rapidly growing, driven by the human-aware robotics research and industry interests. Most approaches require plentiful motion data recorded in diverse environments and settings to train on, as well as for the evaluation \cite{rudenko2020human}. Among the growing number of human trajectory datasets, most focus on capturing interactions between the moving agents in indoor \cite{brscic2013person}, outdoor \cite{robicquet2016learning} and automated driving \cite{bock2020ind} settings. These datasets are designed to study the geometric and velocity aspects of human motion.

Human motion is influenced by a large amount of contextual cues, which include semantic attributes of the static and dynamic environment, space topology and its activity patterns, social roles, relations and preferences of the target agents. Studies of these contextual aspects of human motion are gaining traction, creating the need for new datasets containing relevant cues.

\begin{figure}[!t]
    \centering
    \includegraphics[width=6cm]{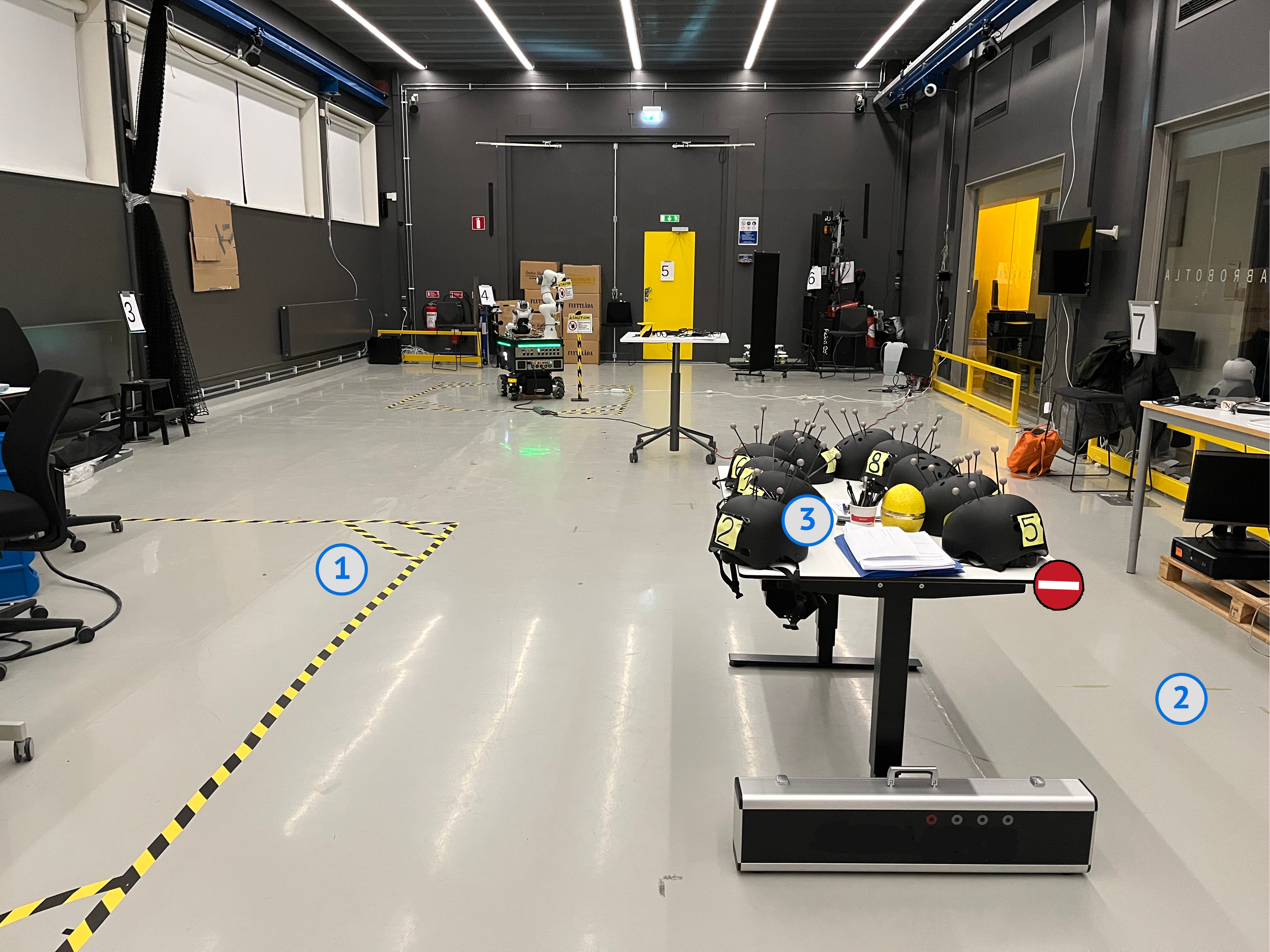}
    \caption{Laboratory room layout including the floor markings (1) in Scenario~1B. The environment contains various static obstacles, including a narrow corridor (2) in the right with entry limited by a no-entry sign. The table displays the motion capture helmets (3).}
    \vspace{-5pt}
    \label{abb:naosetup}
\end{figure}


In this work, we follow on and further develop the TH\"OR protocol for human motion data collection 
introduced in~\cite{rudenko2020thor}. 
There we proposed a weakly-scripted indoor scenario for generating diverse, natural, and goal-driven human motion in crowded social spaces with static obstacles and a moving robot. 
The TH\"OR dataset\footnote{\url{http://thor.oru.se/}}, recorded according to the proposed procedure, includes 9 participants, moving alone and in groups, whose positions and head orientations are tracked with a motion capture system\footnote{\url{https://www.qualisys.com/}}.
The TH\"OR dataset also includes first-person gaze information for a subset of participants. To diversify the recorded motion patterns, participants in TH\"OR move between fixed goal positions in the environment,  receiving at each goal a random card with the next target. The recording features over 60 minutes of motion and over 600 individual and group trajectories. TH\"OR is gaining attention in the scientific community, for instance in robotics~\cite{Zhi2021, Yoon2021} and predictive motion modeling~\cite{Zhi2021_iros}, and serves as a building block for the Atlas motion prediction benchmark \cite{rudenko2022atlas}.

In this paper, we extend THÖR in many aspects. The new recording, which we call \emph{Magni}, includes 160 minutes of motion on 4 acquisition days with a total of 30 unique participants.
In addition to the static obstacles in the room, we 
augment the environment with semantic context,
such as one-way passages and yellow tape markings for areas of caution.

The introduction of the semantic context further enriches the recorded data. 
Moreover, capturing  semantic features enables explainability of motion flow models~\cite{kucner2020probabilistic} or enhances the downstream tasks which require semantics~\cite{rudenko2020semapp}.
To further diversify the recorded motion patterns, in addition to cards indicating the next motion goal of the participants, we introduce remote instructions via voice command (using \textit{Discord}~\cite{discord}). In addition to the gaze directions in the 2D eye-tracker image plane, we also provide 3D 
gaze vectors in the environment reference frame. In addition to the motion capture and eye-gaze data, we record on-board robot sensors (LiDAR, RGB fish-eye, and RGB-D cameras). 
Lastly, we propose two variations in the teleoperated robot motion, namely the ``differential drive'' and ``omnidirectional" motion, which enables the study of human-robot collision avoidance under varying conditions.

This paper presents the data collection procedure, describes sensors, scenarios, and the participants' priming (Sec.~\ref{sec:data_collection}), as well as highlights a portion of the recorded data (Sec.~\ref{sec:recorded_data}). 
We will make the full dataset available in the near future. Once the post-processing is complete, we 
will systematically describe the recorded data and analyze its application in HRI research.

\section{Data Collection}
\label{sec:data_collection}

\subsection{Room Setup}

The room for data collection is the robot lab at Örebro University -- the same as in the THÖR dataset \cite{rudenko2020thor}, which creates continuity between the recordings, while allowing to study human motion in the presence of varying contextual factors and obstacle layouts. Fig.~\ref{fig:room_layout} depicts the room layout. Seven goal positions are placed specifically to drive purposeful navigation through the room, generating frequent interactions between groups in the center. Several static obstacles (robotic manipulators and tables) are placed in the room to prevent walking between goals in a straight path.

Apart from static obstacles, two robots are placed in the room. One is a static robotic arm placed near the podium, as shown on the right in Fig.~\ref{fig:room_layout}. The other one is on the left in Fig.~\ref{fig:room_layout}: an omnidirectional mobile robot with a robotic arm on top (DARKO Robot). In some scenarios, as described in Section~\ref{sec:data_collection:scenarios_description}, the mobile robot is also used for data collection. 
The robot base is RB-Kairos+ and the arm is the Collaborative Robot Panda from Franka Emika. The robot base dimensions are 760$\times$665$\times$690 mm. The maximum reach height of the robot arm is \SI{855}{mm}.
The robot has one Ouster OS0-128 LiDAR, two Azure Kinect RGB-D cameras (one used in these recordings), two Basler fish-eye RGB cameras, and two Sick MicroScan 2D safety LiDARs. The Azure Kinect camera has a 75-degree horizontal field of view and a tracking range of up to \SI{5}{m}. 




   
In one scenario, floor markings to indicate the areas of caution, and stop signs to indicate one-way passages, are added. With black and yellow warning tapes, floor markings are placed around the mobile robot and the robot arm. Two stop signs are placed near the right permanent obstacle, indicating that the passage from right to left is blocked.

\subsection{Scenarios Description}
\label{sec:data_collection:scenarios_description}
We designed three scenarios for diverse data collection, which differ in the room layout, motion mode of the robot and the tasks executed by the participants. In all scenarios, we randomly divided the participants into individuals or groups of two or three people who share the navigation goal. Every group navigates towards their goal point, where it takes a random card, indicating the next goal. Each group takes one card at a time.

Scenario~1 is designed as a baseline to capture
``regular" social behavior of walking people in a 
static environment.
It has two variations: 1A which only includes static obstacles, and 1B which additionally includes floor markings and stop signs in a one-way corridor. 
Scenario~1B is designed with the focus on Maps of Dynamics (MoDs) \cite{kucner2020probabilistic}. MoDs are maps that encode dynamics as a feature of the environment, containing information about motion patterns in an environment. MoDs can provide information for planning and navigation in populated environments. 
The Scenario~1B provides motion data affected by invisible obstacles (areas of caution) and flow controlling signs (one-way passages).

Scenario~2 features the same room layout as Scenario~1A (i.e., without semantics). In addition to the basic goal-driven navigation, this scenario introduces people performing different tasks. These tasks aim to emulate regular activities performed in industrial contexts, such as transporting stacks of different objects between various goal locations. Therefore, in each recording session we assign one participant to carry small objects (i.e., a bucket), and another one to carry medium objects (i.e., a box) between two different goal points. Finally, a group of two people moves a large object (i.e., a poster stand) instructed over \textit{Discord}~\cite{discord}.

   \begin{figure}[t]
      \centering
       \includegraphics[width=7.5cm]{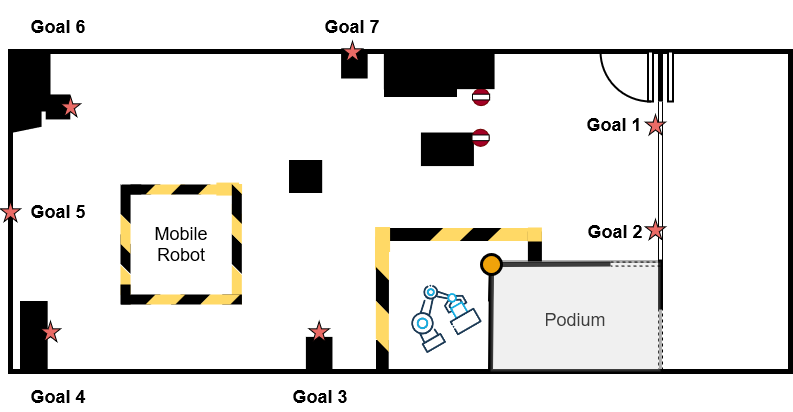}
      \caption{Room layout for Scenario~1B with the focus on Maps of Dynamics. For the other scenarios, we remove the black-yellow striped lane markings. Also, in Scenario~3, the mobile robot on the left becomes a moving obstacle.}
      \label{fig:room_layout}
   \end{figure}

In Scenario~3, the robot (which remained stationary in the previous scenarios, see its position on the left in Fig.~\ref{fig:room_layout}) navigates in the room. Scenario~3 has two variations: 3A, in which the teleoperated robot moves as a regular differential drive robot, and 3B, where the robot moves in an omnidirectional way. In both cases, an operator drives the mobile robot using a remote controller.



\begin{table}[t]
    \begin{center}
    \begin{tabular}{c | c c c}
        \hline
        \textbf{Scenario}  & \textbf{Description} & \textbf{Mobile robot} & \textbf{Duration} \\ \hline
        1A & Baseline motion & Static Obstacle&  8 minutes   \\ \hline
        1B & Semantic features  & Static Obstacle& 8 minutes\\ \hline
        2 & People with tasks & Static Obstacle& 8 minutes \\ \hline
        3A & People with tasks & Directional &  8 minutes \\ \hline
        3B & People with tasks & Omni-directional &  8 minutes \\ \hline
    \end{tabular}
    \end{center}
    \caption{Short description of the conducted scenarios, the motion mode of the robot, and the duration of recordings in one day}
    \vspace{-10pt}
    \label{tab:scenarios}
\end{table}

\subsection{Recording Procedure and Participants' Priming}

\begin{figure}[!t]
\centering
\includegraphics[width=4.2cm]{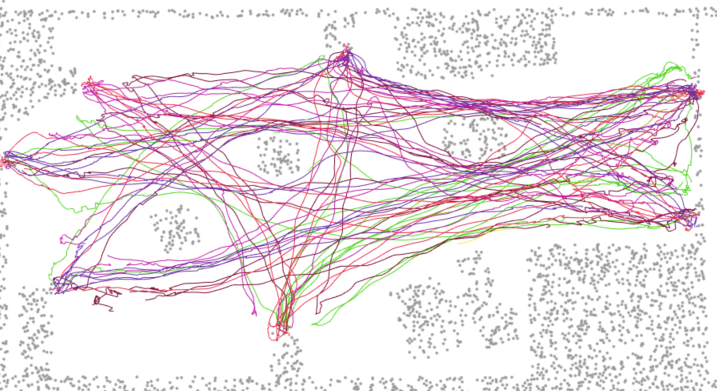}
\includegraphics[width=4.2cm]{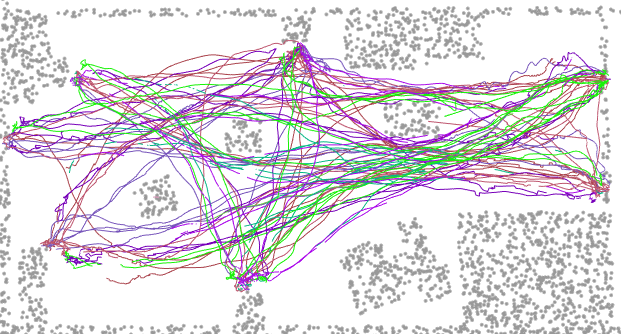}
\caption{Recorded trajectories for one run in Scenario~1A ({\bf left}) and Scenario~1B ({\bf right}), which includes the environment semantics. In both cases, the room contains various static obstacles, including a narrow corridor in the top right area. Trajectories show that most people would instinctively avoid the ``areas of caution'' around the robots, marked with yellow tape (see the layout in Fig.~\ref{fig:room_layout}).}
\vspace{-5pt}
\label{fig:recorded_trajectories}
\end{figure}

\begin{figure}[!t]
\centering
\includegraphics[width=8cm]{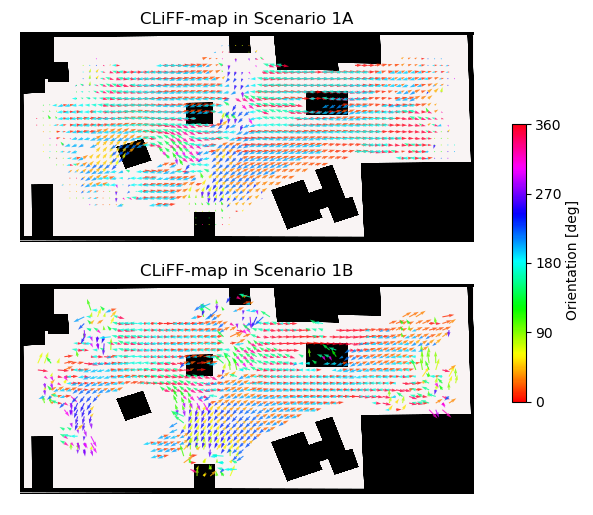}
\caption{Maps of dynamics created from Acquisition I - IV (40 minutes) in Scenario~1A ({\bf top}) and Scenario~1B ({\bf down}). CLiFF-map \cite{kucner2020probabilistic} is used to represent statistical information about flow patterns.}
\vspace{-5pt}
\label{fig:mods}
\end{figure}


At the beginning of each session on the acquisition day, participants filled out a demographic questionnaire. For each scenario variation, on each acquisition day we recorded two runs with a length of 4 minutes. A summary of the scenarios and duration is given in Tab.~\ref{tab:scenarios}. We always started from Scenario~1B to avoid biasing the participants' motion by letting them observe how the lane markings and the stop signs are prepared. After the two runs of this scenario, we followed with Scenario~1A and Scenario~2. Finally, we proceed with each variation from Scenario~3 in no particular order across the recording days.

After each run, participants fill the Raw version of the NASA-Task Load Index (RTLX)~\cite{Hart1988, Hart2006}. The scale consists of a 21-point set of sub-scales [1=Low; 21=High], each of which assesses the mental demand, physical demand, temporal demand, and frustration produced by the task as reported by the participants, as well as their self-perceived performance  and frustration. By the end of the session, after the last run of Scenario~3, participants fill out two extra questionnaires with regard to the mobile robot. First, the Godspeed Questionnaire Series \cite{Bartneck2009}, a semantic differential set of subscales [5-points] that measures the participants' perception of the robot in terms of anthropomorphism, animacy, likeability, perceived intelligence, and perceived safety, respectively. Second, a 5-point likert scale [1=Strongly disagree; 5=Strongly agree] to evaluate trust towards the robot in industrial human-robot collaborations~\cite{charalambous2016}. 
The participants filled out all the questionnaires on paper.


Before recording each run, an instructor calibrates the three eye-trackers (Tobii Glasses 2 and 3) and adjusts the gazes for the Pupil Invisible Glasses. The instructor then returns to the stage and sets a 4-minute alarm. We check with the participants if everyone is ready to begin the measurements. If so, we start the recordings of the motion capture system and the eye trackers simultaneously as the instructor counts down to three to signal the participants the start of a run. Additionally, we record rosbag files including sensor data from the robot platform, like the image feed of its onboard RGB and RGB-D cameras and the point cloud recorded by the LiDAR, as well as topics regarding people tracking. After 4 minutes, we simultaneously stop all recordings and the ringing of the alarm signalizes the participants the end of a run.  

Between each run, while the participants fill out the questionnaires, we prepare the next run; i.e., we remove the floor markings (after the last run of 1A), set up a phone for the \textit{Discord} voice chat (before 2 and 3), check on the batteries of the eye trackers and potentially change them and finally prepare the robots for Scenario~3. As the participants finish filling out the questionnaire, we shuffle the roles in Scenario~2 and 3 and always assign new groups consisting out of one to three participants for the next run, hereby we always follow the rule, that for groups there can only be one participant with an eye tracker. We assign each group a new goal point to start from at the next run. For the scenarios with roles (2 and 3) we also give a short recap on the task connected with each role, if that participant has not been assigned this role before.

\section{Recorded Data}
\label{sec:recorded_data}


We recorded data on 4 acquisition days for a total of 30 unique participants (9 on Day I, 7 on Days II-IV).
As described in Sec.~\ref{sec:data_collection:scenarios_description}, each acquisition day consists of three different scenarios, and two of them have two different variants. Furthermore, we recorded two 4-minute runs per scenario. Therefore, each acquisition comprises ten runs comprehending all scenarios and yielding 40 minutes of multi-modal data: 3D motion patterns, eye-gaze data from 3 eye trackers, and robot sensor data.

Fig.~\ref{fig:recorded_trajectories} shows 2D motion trajectories, collected during one 4-minute run in Scenario~1A (left) and Scenario~1B (right). It shows the difference between the two scenarios in areas delimited by the lane markings (see Fig. 2 for the layout reference). Specifically, participants in Scenario~1B tended to navigate farther from the delimited static objects than in Scenario~1A. In addition, Fig.~\ref{fig:mods} shows the maps of dynamics~\cite{kucner2020probabilistic} generated from the collected trajectories from all runs in Scenario~1A and 1B. It shows that in Scenario~1B the flow is less intensive near the ``areas of caution" around the robots. Also, one-way passage flow pattern in the top right corner from Scenario~1B is clearly visible.

\begin{figure}[!t]
\centering
\includegraphics[width=8.3cm]{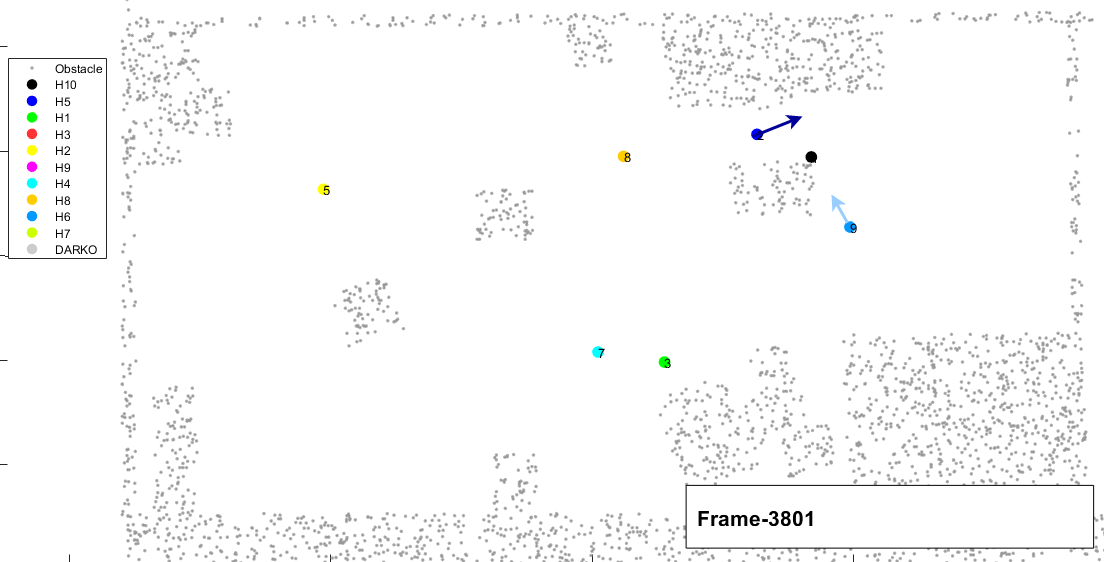} \\
\vspace{3pt}
\includegraphics[width=4.2cm]{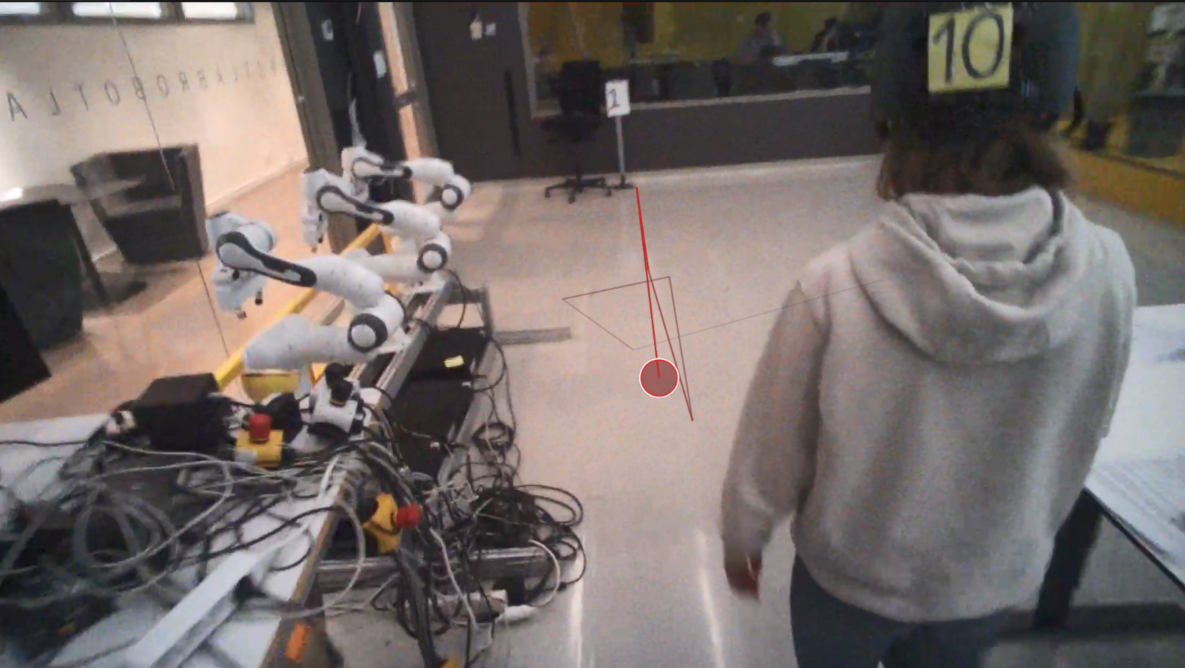}
\includegraphics[width=4.2cm]{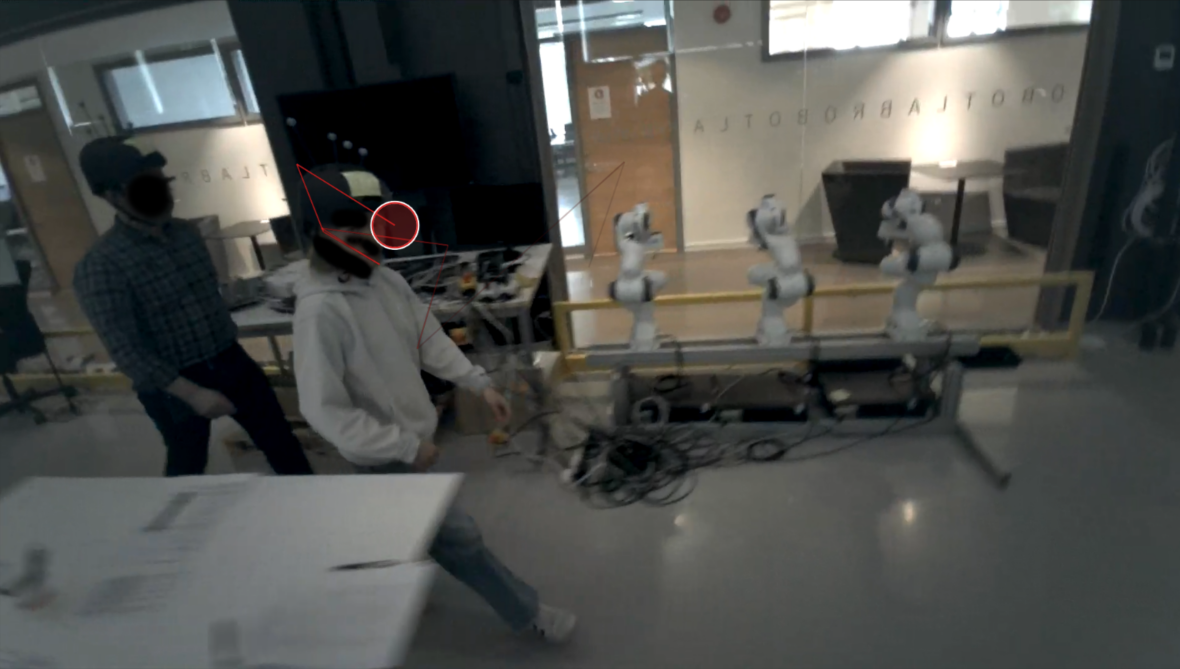}
\caption{Eye-gaze vectors, recorded for the participants wearing eye-tracking glasses. {\bf Top:} gaze-vectors mapped into the 3D global map frame for participants 2 (dark blue arrow) and 9 (light blue arrow). {\bf Bottom:} corresponding first-person views for participants 2 and 9. Red line displays the gaze history in the past 2 seconds, and the red circle shows the gaze point in the current frame.}
\vspace{-5pt}
\label{fig:eye_gazes}
\end{figure}

\begin{figure}[t]
  \centering
   \includegraphics[width=7cm]{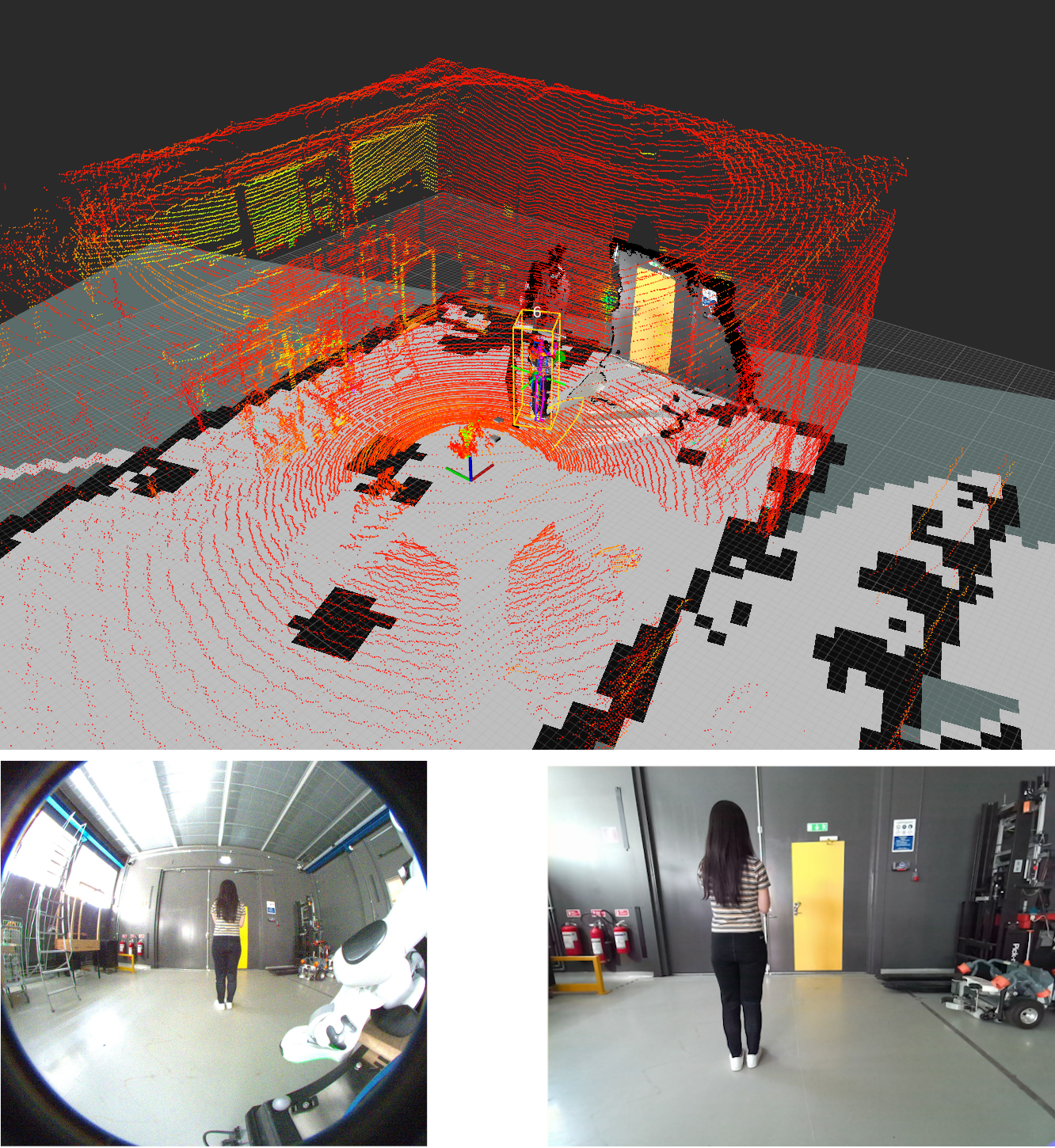}
    \caption{Data collected from the moving robot. \textbf{Top:} 3D visualization of sensor data in RViz: Ouster point cloud shown as red/yellow points, an occupancy grid map for the laboratory and the output of the people tracking module (yellow bounding box) \cite{linder2020accurate}. \textbf{Bottom left:} Fish-eye RGB camera image. \textbf{Bottom Right:} Azure Kinect RGB-D camera image.}
    \vspace{-5pt}
  \label{fig:darko_rviz}
\end{figure}

Fig.~\ref{fig:eye_gazes} shows the example eye-gazes, recorded for two participants wearing the tracking glasses in the same frame. The 2D gaze direction is provided in the first-person video frame, furthermore we calculate the 3D gaze coordinates in global map frame. Finally, Fig.~\ref{fig:darko_rviz} provides an example of the data recorded with the on-board robot sensors (LiDAR, RGB and fish-eye cameras), displayed in RViz.




\section{Conclusion and Future Work}

In this paper we present a new contextually-rich recording of human-robot co-navigation in an indoor environment. The multi-modal data on human motion, collected from the motion capture system, eye-gaze trackers and the on-board sensors of a moving robot, aims to supply the research on human motion prediction, obstacle avoidance, maps of dynamics and human-robot interaction.

In future work we plan to extend the co-navigation scenarios with explicit forms of human-robot communication, for instance by signalling the robot's intentions, and collaboration, for instance in loading, transporting and unloading the boxes.

\bibliographystyle{IEEEtran}
\bibliography{IEEEabrv,references}

\end{document}